\documentclass[sigconf]{acmart}
\AtBeginDocument{%
  \providecommand\BibTeX{{%
    \normalfont B\kern-0.5em{\scshape i\kern-0.25em b}\kern-0.8em\TeX}}}

\copyrightyear{2022}
\acmYear{2022}
\setcopyright{acmcopyright}\acmConference[WWW '22 Companion]{Companion Proceedings of the Web Conference 2022}{April 25--29, 2022}{Virtual Event, Lyon, France.}
\acmBooktitle{Companion Proceedings of the Web Conference 2022 (WWW '22 Companion), April 25--29, 2022, Virtual Event, Lyon, France}
\acmPrice{15.00}
\acmDOI{10.1145/3487553.3524637}
\acmISBN{978-1-4503-9130-6/22/04}
\begin{document}

\title{FinRED: A Dataset for Relation Extraction in Financial Domain}

\author{Soumya Sharma}
% \authornote{Main correspondent}
\email{soumyasharma20@gmail.com}
\affiliation{%
  \institution{IIT Kharagpur}
  \country{India}
}

\author{Tapas Nayak}
% \authornote{This work was done when the author was a postdoctoral researcher at IIT Kharagpur.}
\email{tnk02.05@gmail.com}
\affiliation{%
  \institution{IIT Kharagpur}
  \country{India}
}

\author{Arusarka Bose *}
\thanks{* Equal contribution}
\email{arusarkabose@gmail.com}
\affiliation{%
  \institution{IIT Kharagpur}
  \country{India}
}

\author{Ajay Kumar Meena *}
\email{ajaymeena3210@gmail.com}
\affiliation{%
  \institution{IIT Kharagpur}
  \country{India}
}

\author{Koustuv Dasgupta}
\email{Koustuv.x.Dasgupta@gs.com}
\affiliation{%
  \institution{Goldman Sachs}
  \country{India}
}

\author{Niloy Ganguly}
\email{niloy@cse.iitkgp.ac.in}
\affiliation{%
  \institution{IIT Kharagpur, India}
  \country{LU, Hannover} %Germany}
}

\author{Pawan Goyal}
\email{pawang@cse.iitkgp.ac.in}
\affiliation{%
  \institution{IIT Kharagpur}
  \country{India}
}

\renewcommand{\shortauthors}{Soumya Sharma et al.}

\begin{abstract}
  Relation extraction models trained on a source domain cannot be applied on a different target domain due to the mismatch between relation sets. In the current literature, there is no extensive open-source relation extraction dataset specific to the finance domain. In this paper, we release FinRED, a relation extraction dataset curated from financial news and earning call transcripts containing relations from the finance domain. FinRED has been created by mapping Wikidata triplets using distance supervision method. We manually annotate the test data to ensure proper evaluation. We also experiment with various state-of-the-art relation extraction models on this dataset to create the benchmark. We see a significant drop in their performance on FinRED compared to the general relation extraction datasets which tells that we need better models for financial relation extraction.
\end{abstract}





\maketitle

\section{Introduction}

The task of relation extraction (RE) is defined as identifying triplets from text. Recently joint entity and relation extraction models \cite{nayak2020effective,sui2020joint,wang-etal-2020-tplinker} have been proposed to eliminate the dependency on an external NER module. These models have achieved remarkable performance in this task. Freebase-New York Times dataset (FB-NYT) \cite{hoffmann-etal-2011-knowledge} is obtained by mapping relational triplets from Freebase \cite{freebase}, WebNLG \cite{gardent-etal-2017-creating} is a natural language generation dataset created semi-automatically from DBPedia~\cite{Bizer2009DBpediaA} triplets. Although KBs such as Freebase, Wikidata, DBpedia contain a lot of relational triplets from different domains such as finance, material science, legal, but that is not reflected in the existing RE datasets as they  very generic sources of text corpus.  There have been some efforts to create datasets for particular domain-specific relation extraction such as legal \cite{andrew-2018-automatic}, biomedical \cite{Gu2016ChemicalinducedDR,Li2017ANJ,Choi2018ExtractionOP,thillaisundaram-togia-2019-biomedical}, scientific articles \cite{luan-etal-2017-scientific,jain-etal-2020-scirex}. There have been some efforts to create datasets in the financial domain \cite{vela-declerck-2009-concept, Wu2020CreatingAL,jabbari-etal-2020-french}, however to the best of our knowledge, there is currently no standard dataset in English to understand financial relations. There has been some work on financial numeral understanding \cite{chen2019overview,chen-etal-2019-numeracy}, however, our paper focuses on textual relations and not numerical.
To bridge this gap, we create a {\bf finance domain-specific distance supervised relation extraction dataset}. We obtain the finance relations from Wikidata KB and use text from the finance domain to create the \textit{FinRED} dataset.

\section{Dataset Sources and Creation}
The FinRED dataset \footnote{Repository\: \href{https://github.com/soummyaah/FinRED/}{https://github.com/soummyaah/FinRED/}} has been created using two sets of documents: 1) Webhose Financial News, and 2) Earning Call Transcript (ECT).

\textbf{Webhose Financial News: } For the financial news articles, we use the freely available dataset curated by Webhose. It contains 47,851 English financial news articles crawled from July 2015 - October 2015.

\textbf{Earning Call Transcripts (ECT): } We collect 4,713 ECTs dated from June 2019 to September 2019 from seekingalpha.com. A typical ECT contains a presentation from the company participants followed by a questionnaire section containing questions asked by audience answered by company participants. We term the presentation portion and the answers as monologues. We use the presentation as well as the questionnaire portion of the transcript and remove monologues with less than 200 characters. 
%Here monologue refers to a monologue given by a company participant during the presentation or the response given by a company participant during the questionnaire. 
In the ECT corpus, we have about 200K monologues, 1.8M sentences with an average of 7.19 sentences per monologue. In total, we have 152K tokens in the corpus.

% We present a few statistics for the earning call transcript corpus in Table ~\ref{tab:ect}. 

% \begin{table}[ht]
% \centering
% \small
% \caption{Details of the Earning Call Transcript data collected in terms of property-value pairs}
% \label{tab:ect}
% \begin{tabular}{ |l|c|} 
% \hline
% \textbf{Property} & \textbf{Value} \\ \hline
% No. of Transcripts & 4,713 \\ \hline
% % No. of companies & \\ \hline
% No. of monologues & 2,59,951 \\ \hline
% No. of sentences & 18,70,232 \\ \hline
% Average No. of sentences per monologue & 7.19 \\ \hline
% Average No. of words per sentence & 20.67 \\ \hline
% No. of tokens & 1,52,239 \\ \hline
% \end{tabular}
% \end{table}
% \begin{figure}[ht]
% \centering
% \includegraphics[scale=0.6]{ECT_doc.drawio (1).png}
% \caption{Typical structure of an Earning Call Transcript}
% \label{fig:ect}
% \end{figure}
\begin{table}[ht]
\centering
\small
\caption{Comparison between different RE datasets.}
\label{tab:comparing_datasets}
\begin{tabular}{ |l|c|c|c|c|c|c|} 
\hline
% Relation & \#Triplet & Relation & \#Triplet & Relation & \#Triplet \\ \hline
Dataset Name  & Train & Test & \#Relations & \#Financial Relations \\ \hline
 FB-NYT  & 56,196 & 5,000 & 24 & 4 \\ \hline 
 WebNLG  & 5,519 & 703  & 216 & 12 \\ \hline
%  DocRED & Document & 4,053 & 1,000 & 96 & 33\\ \hline
 \textbf{FinRED (Ours)}  & \textbf{5,699} & \textbf{1,068} & \textbf{29} & \textbf{29} \\ \hline

\end{tabular}
\end{table}

\begin{table}[ht]
\centering
\small
\caption{Examples in the dataset}
\label{tab:annotation_example}
\begin{tabular}{ |c||c|c|c|} 
\hline
\textbf{Sentence} & \textbf{Head} & \textbf{Relation} & \textbf{Tail} \\ \hline
Antony Jenkins has been   & & chief & \\
sacked as chief executive   & Anthony &  executive & \\
officer of Barclays Plc. &  Jenkins & officer & Barclays \\ \hline
MEXICO CITY — State-owned & & product or & \\ Mexican oil company Pemex & & material & \\
is reporting second quarter & Pemex & produced & petroleum \\
losses of $\$$US5.2 billion & & & \\
($\$$A7.16 billion) due mainly & & headquarters & Mexico \\
to lower petroleum prices & Pemex & location & City \\ \hline
\end{tabular}
\end{table}
\textbf{Knowledge Base (KB): } We use a subset of the Wikidata KB as a source of relational triplets for financial domain containing manually filtered 29 financial relations. Using the idea of distance supervision \cite{mintz-etal-2009-distant}, we align the relational triplets to the text corpus.

\textbf{Dataset: } In total we obtain about 21,000 sentences using the distance supervision method, however a lot of these sentences are noisy. Eg: ``Delhi-based National Housing Bank (NHB) is working to set up more than 80 new housing finance companies (HFCs), with a special emphasis on those that will focus on financing affordable houses." where the tuple is (More Than, industry, finance) since in Wikidata, ``More Than`` is listed as an insurance company of the financial industry. We study a small representation of the data in detail and remove datapoints with incorrect entities such as ``More Than`` and ``industry``. After reduction, we get 7,775 sentences which we divide into the train, dev, and test dataset. We observe that this dataset only contains 920 sentences from the earnings call transcript and we attribute this to the earning call transcripts containing primarily conversational data which often does not contain a triplet in a sentence. A comparison of \textit{FinRED} with FB-NYT \cite{hoffmann-etal-2011-knowledge} and WebNLG \cite{gardent-etal-2017-creating} along with a few examples from the dataset has been showcased in Table ~\ref{tab:comparing_datasets}.

\textbf{Annotation of test data: } We annotate the dataset with the help of 2 non-native fluent English speakers annotators, who go through the dataset twice, and all incorrect triplets are removed. All triplets marked as incorrect by both the annotators are removed from the dataset. The Cohen $\kappa$ between the annotators is 82.1\%. 

\section{Experiments}

To avoid the use of external NER module, we choose 3 joint entity and relation extraction models, namely, SPN \cite{sui2020joint}, TPLinker \cite{wang-etal-2020-tplinker}, and CasRel \cite{wei2020novel} for our experiments. These models achieved state-of-the-art performance on standard relation extraction datasets Freebase-New York Times \cite{hoffmann-etal-2011-knowledge} and WebNLG \cite{gardent-etal-2017-creating}. We report precision, recall and F1 score for triplet extraction based on exact entity matching criterion and report the results in Table \ref{tab:main_results}.

From Table \ref{tab:main_results}, we can observe a drop of $\sim$4\% for SPN and $\sim$25-30\% for TPLinker and Casrel models in F1 as compared to FB-NYT and WebNLG datasets. Drop in the performance showcases that more research is required to better capture performance. For the SPN model, we observe that the relation F1 value is 88.37\% and the entity F1 value is 96.36\% showcasing that while the model can predict the entity properly, the lower model performance can be attributed to incorrect relation classification. We see similar trends in the performance of TPLinker and CasRel. 

\begin{table}[ht]
\centering
\caption{Performance of the three baseline models on FinRED.}
\label{tab:main_results}
\begin{tabular}{|l||c|c|c|c|c|}
\hline
         & \multicolumn{3}{c|}{FinRED (Ours)} & FB-NYT & WebNLG \\ \hline
Model    & Prec.     & Rec.     & F1   & F1        & F1     \\ \hline
SPN      & \textbf{89.22}\% & \textbf{86.42}\% & \textbf{87.80}\%  & 92.30\%     & 91.80\% \\ \hline
TPLinker & 76.59\% & 63.45\% & 69.40\% & 92.00\%     &  86.70\%   \\ \hline
CasRel   & 69.71\% & 55.84\% & 62.01\% &  89.60\%   &  93.40\%  \\ \hline
\end{tabular}
\vspace{-3mm}
\end{table}

\begin{table}[ht]
\centering
\small
\caption{Performance of SPN model on the sentence classes with different number triplets and  with type of overlapping triplets in them.}
\label{tab:triplet_class}
\begin{tabular}{|l|l|l|l|}
    \hline
        \textbf{Metric} & \textbf{Precision} & \textbf{Recall} & \textbf{F1}  \\ \hline
        \textbf{1 triplet} & 76.15\% & 86.46\% & 80.98\%  \\ \hline
        \textbf{2 triplets} & 86.36\% & 88.88\% & 87.60\%  \\ \hline
        \textbf{3 triplets} & 93.31\% & 85.24\% & 89.09\%  \\ \hline
        \textbf{4+ triplets} & 95.79\% & 88.61\% & 92.06\%  \\ \hline
        % \textbf{5 triplets+} & 95.89\% & 81.30\% & 88\%  \\ \hline
        \textbf{NEO} & 78.40\% & 89.61\% & 83.36\%  \\ \hline
        \textbf{EPO} & 93.07\% & 85.50\% & 89.16\%  \\ \hline
        \textbf{SEO} & 89.86\% & 86.18\% & 87.98\% \\ \hline
% Rel F1 & 81.33\% & 88.76\% & 89.36\% & 92.35\% & 88.94\% \\ \hline
% Entity F1 & 100\% & 99.13\% & 94.94\% & 95.52\% & 90.82\% \\ \hline
\end{tabular}
\end{table}
Performance of SPN model on the sentence classes with different number of triplets is reported in Table ~\ref{tab:triplet_class}. The performance improves as the number of triplets increases which showcases that the more the number of triplets( typically EPO and SEO) , the better assistance the machine has to make better predictions.
% . We hypothesise that a few more triplets (typically EPO and SEO) assist the machine make better predictions, but too many confuse it.

In Table ~\ref{tab:triplet_class} we also report the performance of SPN model on the sentence classes with the type of overlapping triplets in them. Based on the overlap of entities, the sentence can be divided into three types: 1) NEO (No entity overlap) 2) EPO (Entity Pair Overlap) and 3) SEO (Single Entity Overlap). Here, we can observe that the model shows a slightly higher performance for SEO (F1: 87.98\%) and EPO (F1: 89.16\%) as compared to NEO (F1: 83.36\%). We surmise that this is because the entity prediction part of the model performs well so for multi-label and overlapping triplets, the performance is higher since it needs to identify less number of entities.

\section{Conclusion}
In this paper, we propose \textit{FinRED}, a relation extraction dataset for the finance domain curated from earning call transcript corpus and financial news articles. It contains more finance domain-specific relations than the existing RE datasets. We experimented with three state-of-the-art joint entity and relation extraction models on this dataset and saw a significant drop in F1 score compare to general domain RE datasets showcasing that more research is required on the models for FinRED.

\begin{acks}
This research was partially supported by Goldman Sachs sponsored research grant FTHS. (FinTalk: Research towards creating a platform for highlight generation and summarization of financial documents while taking into account user feedback).
\end{acks}
\bibliographystyle{ACM-Reference-Format}
\bibliography{sample-base}

% \appendix









\end{document}